\documentclass{article} % For LaTeX2e
\usepackage{iclr2017_conference,times}

\usepackage{booktabs}       % professional-quality tables
\usepackage{amsfonts}       % blackboard math symbols
\usepackage{nicefrac}       % compact symbols for 1/2, etc.
\usepackage{microtype}      % microtypography
\usepackage{setspace}

\usepackage{qiangstyle}
\usepackage{url}
\newcommand{\dilinfig}{./figures}
\newcommand{\dilincheck}[1]{#1}%\textcolor{magenta}{#1}}

\title{Learning to Draw Samples: With Application to Amortized MLE for Generative Adversarial Learning} %Training Deep Energy Models}

\author{Dilin Wang, ~~Qiang Liu\\
Department of Computer Science, Dartmouth College\\
\texttt{\{dilin.wang.gr, qiang.liu\}@dartmouth.edu}
}
\begin{document}

\maketitle

\begin{abstract}
We propose a simple algorithm to train stochastic 
neural networks 
%black-box procedures, such as neural networks with random inputs, 
to draw samples from given target distributions for probabilistic inference. 
Our method is based on iteratively adjusting the neural network parameters 
so that the output changes along a Stein variational gradient \citep{liu2016stein} that maximumly decreases 
the KL divergence with the target distribution. 
%to decrease the KL divergence between the output random variable and the target distribution, by mimicing a recently proposed Stein variational gradient descent. 
Our method works for any target distribution specified by their unnormalized density function, 
and can train any black-box architectures that are differentiable in terms of the parameters we want to adapt. 
%By allowing to ``learn to draw samples'', our method opens a host of applications. 
%We present two examples in this paper: 
%In addition, we leverage our method to 
As an application of our method, we propose an  \emph{amortized MLE} algorithm for training deep energy model, where a neural sampler is adaptively trained to 
approximate the likelihood function. Our method mimics an adversarial game 
between the deep energy model and the neural sampler, and obtains realistic-looking images competitive with the state-of-the-art results. 
% that attempts to fool the energy model.
%by training neural samplers to approximate the gradient for maximum likelihood by training neural networks to approximate the derivative of the log partition function of unnormalized distributions
%for maximum likelihood training, % models,  
%we propose an \emph{amortized MLE} method for training deep energy model, 
%which admits an adversarial game between 
%turning the traditional MLE algorithm into a generative adversarial  
%between the deep energy model 
%and the generative neural network that attempts to fool the energy model. We show that we can get realistic-looking images competitive with the state-of-the-art results. 
%2) by treating stochastic gradient Langevin dynamics as a black-box sampler, we train to automatically adjust its learning rate to maximize its convergence speed, outperforming the hand-designed learning rate schemes. % such as adagrad. 
%generative adversarial networks. 
\end{abstract}

\section{Introduction}

Modern machine learning increasingly relies on highly complex probabilistic models to reason about uncertainty.  
A key computational challenge is to develop efficient inference techniques to approximate, or draw samples from complex distributions. 
Currently, most inference methods, including MCMC and variational inference, are hand-designed by researchers or domain experts. 
This makes it difficult to fully optimize the choice of different methods and their parameters, and exploit the structures in the problems of interest in an automatic way. 
The hand-designed algorithm can also be inefficient when it requires to make fast inference repeatedly on a large number of different distributions with similar structures. 
This happens, for example, when we need to reason about a number of observed datasets in settings like online learning, 
or need fast inference as inner loops for other algorithms such as maximum likelihood training. 
Therefore, it is highly desirable to develop more intelligent probabilistic inference systems that can adaptively improve its own performance to fully the optimize computational efficiency, and generalize to new tasks with similar structures. 

Specifically, denote by $p(x)$ a probability density of interest specified up to the normalization constant, which we want to draw sample from, or marginalize to estimate its normalization constant. We want to study the following problem: 

\begin{problem}\label{pro:prob1}
Given a distribution with density $p(x)$ %specified up to the normalization constant, 
and a function $f(\eta;~\xi)$ with parameter $\eta$ and random input $\xi$, 
for which we only have assess to draws of the random input $\xi$ (without knowing its true distribution $q_0$), 
and the output values of $f(\eta;~\xi)$ and its derivative $\partial_\eta f(\eta;~\xi)$ given $\eta$ and $\xi$.  
We want to find an optimal parameter $\eta$ so that the density of the random output variable $x = f(\eta;~\xi)$ with $\xi\sim q_0$ closely matches the target density $p(x)$. 
\end{problem}

Because we have no assumption on the structure of $f(\eta;~\xi)$ 
and the distribution of random input,
we can not directly calculate the actual distribution of the output random variable $x = f(\eta;~\xi)$; 
this makes it difficult to solve Problem~\ref{pro:prob1} using the traditional variational inference (VI) methods. 
Recall that traditional VI approximates $p(x)$ using simple proposal distributions $q_\eta(x)$ indexed by parameter $\eta$, 
and finds the optimal $\eta$ by minimizing KL divergence $\KL(q_\eta ~||~p) = \E_{q_\eta}[\log (q_\eta/p)]$, 
which requires to calculate the density $q_\eta(x)$ or its derivative that is not computable by our assumption 
(even when the Monte Carlo gradient estimation and the reparametrization trick \citep{kingma2013auto} are applied).  

In fact, it is this requirement of calculating $q_\eta(x)$ 
that has been the major constraint for the designing of state-of-the-art variational inference methods with rich approximation families; 
the recent successful algorithms \citep[e.g.,][to name only a few]{rezende2015variational,tran2015variational,ranganath2015hierarchical} 
have to handcraft special variational families 
to ensure the computational tractability of $q_\eta(x)$ and simultaneously obtain high approximation accuracy, 
which require substantial mathematical insights and research effects. 
Methods that do not require to explicitly calculate $q_\eta(x)$ 
can significantly simplify the design and applications of VI methods, 
allowing practical users to focus more on choosing proposals that work best with their specific tasks. %, making the design and applications of VI much easier. 
%We should distinguish 
We will use the term \emph{wild variational inference} to refer to new variants of variational methods that require no tractability $q_\eta(x)$, %\emph{wild variational inference}, 
to distinguish with the \emph{black-box variational inference} \citep{ranganath2013black}
which refers to methods that work for generic target distributions $p(x)$ without significant model-by-model consideration (but still require to calculate the proposal density $q_\eta(x)$). 

A similar problem also appears in importance sampling (IS), 
where it requires to calculate the IS proposal density $q(x)$ in order to calculate the importance weight $w(x) = p(x)/q(x)$. 
However, %it is been shown that it is possible to develop %\emph{black-box importance sampling}  
there exist methods that use no explicit information of $q(x)$, which, seemingly counter-intuitively, give better asymptotic variance or converge rates than the typical IS that uses the proposal information \citep[e.g.,][]{liu2016black, briol2015probabilistic, henmi2007importance, delyon2014integral}.
% at least in the asymptotic sense. 
%This phoemnomin was first observed 
Discussions on this phenomenon dates back to \citet{o1987monte}, who 
argued that ``Monte Carlo (that uses the proposal information) is fundamentally unsound'' for violating the Likelihood Principle, 
and developed Bayesian Monte Carlo \citep{o1991bayes} as 
an example that uses no information on $q(x)$, yet gives better convergence rate than the typical Monte Carlo $\Od(n^{-1/2})$ rate \citep{briol2015probabilistic}. 
Despite the substantial difference between IS and VI, 
these results intuitively suggest the possibility of developing efficient variational inference without calculating $q(x)$ explicitly. 

In this work, we propose a simple algorithm for Problem~\ref{pro:prob1} 
by iteratively adjusting the network parameter $\eta$ to make its output random variable changes along 
a Stein variational gradient direction (SVGD) \citep{liu2016stein} that optimally decreases its KL divergence 
with the target distribution. 
Critically, the SVGD gradient includes a repulsive term to ensure that the generated samples have the right amount of variability that matches $p(x).$ 
In this way, we ``amortize SVGD'' using a neural network, which makes it possible for our method to adaptively improve its own efficiency  
%ccording to which shared inferences are cached and composed together to answer new queries. 
by leveraging fast experience, especially in cases when it needs to perform fast inference repeatedly on a large number of similar tasks. 
%One example of this is in maximum likelihood estimation (MLE) for unnormalized distributions, in which it is required to draw samples to approximate the likelihood at each iteration. 
%referred to as
As an application, we use our method to amortize the MLE training of deep energy models, where a neural sampler is adaptively trained to 
approximate the likelihood function. Our method, which we call \emph{SteinGAN}, mimics an adversarial game 
between the energy model and the neural sampler, and obtains realistic-looking images competitive with the state-of-the-art results produced by generative adversarial networks (GAN) \citep{goodfellow2014generative, radford2015unsupervised}. 

\paragraph {Related Work} 
The idea of amortized inference \citep{gershman2014amortized}
 has been recently applied in various domains of probabilistic reasoning, 
 including both amortized variational inference \citep[e.g.,][]{kingma2013auto, jimenez2015variational}, 
and data-driven proposals for (sequential) Monte Carlo methods \citep[e.g.,][]{paige2016inference}, 
 %and maximum a posteriori \citep{sonderby2016amortised}, 
 to name only a few. Most of these methods, however, 
 require to explicitly calculate $q(x)$ (or its gradient). 
 One exception is a very recent paper \citep{operator} that avoids calculating $q(x)$ using an idea related to Stein discrepancy \citep{gorham2015measuring, liu2016kernelized, oates2014control, chwialkowski2016kernel}.
 %; the method by \citet{operator} may be more computationally expensive than our method because it requires to train an additional neural network to distinguish $q_\eta$ and $p$ and requires to calculate the second derivative of $p(x)$. 
 There is also a raising interest recently on a similar problem of ``learning to optimize'' \citep[e.g.,][]{andrychowicz2016learning, daniel2016learning, li2016learning}, which is technically easier than the more general problem of ``learning to sample''. %that also requires to take the uncertainty into account.  
In fact, we show that our algorithm reduces to ``learning to optimize'' when only one particle is used in SVGD. 

Generative adversarial network (GAN) and its variants have recently gained remarkable success on generating realistic-looking images 
\citep{goodfellow2014generative, salimans2016improved, radford2015unsupervised, li2015generative,dziugaite2015training,nowozin2016f}.  
All these methods are set up to train latent variable models (the generator) under the assistant of the discriminator.  
Our SteinGAN instead performs traditional MLE training for a deep energy model, 
 with the help of a neural sampler that learns to draw samples from the energy model to approximate the likelihood function; 
 this admits an adversarial interpretation: we can view the neural sampler as a generator that attends to fool the deep energy model, which in turn serves as a discriminator 
that distinguishes the real samples and the simulated samples given by the neural sampler. 
This idea of training MLE with neural samplers was first discussed by \citet{kim2016deep}; 
one of the key differences is that the neural sampler in \citet{kim2016deep} 
is trained with the help of a heuristic diversity regularizer based on batch normalization, while 
SVGD enforces the diversity in a more principled way. 
%which, without a principled approach for solving Problem~\ref{pro:prob1},  in which the diversity of simulated samples are enforced with a heuristic motivated by batch normalization.  
Another method by \citet{zhao2016energy} also trains an energy score to distinguish real and simulated samples, but within a non-probabilistic framework (see Section~\ref{sec:gan} for more discussion). 
Other more traditional approaches for training energy-based models  \citep[e.g.,][]{ngiam2011learning, xie2016theory} are often based on variants of MCMC-MLE or contrastive divergence \citep{geyer1991markov, hinton2002training, tieleman2008training}, and have difficulty generating realistic-looking images from scratch.

\section{Stein Variational Gradient Descent (SVGD)}
%This proposal will focus on integrating Stein's method with variational inference, by leveraging a new variational interpretation of Stein's characterization \eqref{equ:stein00} discovered in my recent work \citep{liu2016stein}, in which a general purpose inference algorithm called Stein variational gradient descent (SVGD) has been proposed. 
Stein variational gradient descent (SVGD) \citep{liu2016stein} is a general purpose Bayesian inference algorithm motivated by 
Stein's method \citep{stein1972, barbour2005introduction} and kernelized Stein discrepancy \citep{liu2016kernelized, chwialkowski2016kernel, oates2014control}. 
It uses an efficient \emph{deterministic} gradient-based update 
to iteratively evolve a set of particles $\{x_i\}_{i=1}^n$ to minimize the KL divergence with the target distribution. 
SVGD has a simple form that reduces to the typical gradient descent for maximizing $\log p$ when using only one particle $(n=1)$, 
and hence can be easily combined with the successful tricks for gradient optimization, 
including stochastic gradient, adaptive learning rates (such as adagrad), and momentum. % making large scale Bayesian inference both much easier and more efficient. 

To give a quick overview of the main idea of SVGD, let $p(x)$ be a positive density function on $\R^d$ which we want to 
approximate with a set of particles 
$\{ x_i\}_{i=1}^n$. 
SVGD initializes the particles by sampling from some simple distribution $q_0$, and 
updates the particles iteratively by 
%performs iterative updates of form 
\begin{align}\label{equ:xxii}
x_i  \gets x_i +  \epsilon \ff(x_i),  ~~~~ \forall i = 1, \ldots, n,  
\end{align}
where $\epsilon$ is a step size, and 
$\ff(x)$ is a ``particle gradient direction'' %  perturbation direction, or velocity field, 
%which roughly speak, is the gradient of $\KL(q_\vx~||~ p)$ w.r.t. $\vx$, 
chosen to maximumly decrease the KL divergence between the distribution of particles and the target distribution, in the sense that  
%decreases with the fastest speed in the sense that 
\begin{align}\label{equ:ff00}
\ff =   \argmax_{\ff \in \F}  \bigg\{  -   \frac{d}{d\epsilon} \KL(q_{[\epsilon\ff]} ~|| ~ p) \big |_{\epsilon = 0}  \bigg\}, 
%\frac{1}{\epsilon}\{ \KL(q_{t+1}~||~ p )  -  \KL(q_{t}~||~ p) \}, 
\end{align}
where $q_{[\epsilon \ff]}$ denotes the density of the updated particle $x^\prime = x + \epsilon \ff(x) $ when the density of the original particle $x$ is $q$, and $\F$ is the set of perturbation directions that we optimize over. 
We choose $\F$ to be the unit ball of a vector-valued reproducing kernel Hilbert space (RKHS) $\H^d = \H \times \cdots \times \H$ with each $\H$ associating with a positive definite kernel $k(x,x')$; 
%this makes $\F$
%this choice of $\F$ is broad, since $\H^d$
note that $\H$ is dense in the space of continuous functions with universal kernels such as the Gaussian RBF kernel. 

Critically, the gradient of KL divergence in \eqref{equ:ff00} 
equals a simple linear functional of $\ff$, allowing us to obtain a closed form solution for the optimal $\ff$. 
\citet{liu2016stein} showed that 
%A key observation is that the objective in \eqref{equ:ff00} is a linear functional of $\ff$, in fact, we have
\begin{align}\label{equ:klstein00}
&- \frac{d}{d\epsilon} \KL(q_{[\epsilon\ff]} ~|| ~ p) \big |_{\epsilon = 0}  = \E_{x\sim q}[\sumstein_p \ff(x)], \\[.5\baselineskip]
&~~~~\text{with}~~~~~ \sumstein_p \ff(x)  = \nabla_x \log p(x) ^\top \ff (x)+ \nabla_x \cdot \ff(x),  
\end{align}
where $\sumstein_p$ is considered as a linear operator acting on function $\ff$ and is called the Stein operator in connection with Stein's identity which shows that 
the RHS of \eqref{equ:klstein00} equals zero if $p = q$: 
\begin{align}\label{equ:steinid}
\E_{p}[\sumstein_p \ff] =\E_{p}[ \nabla_x \log p ^\top \ff + \nabla_x \cdot \ff] = 0. 
\end{align}
This is a result of integration by parts assuming the value of $p(x)\ff(x)$ vanishes on the boundary of the integration domain.   
%Obviously, $\F$ should be taken as broad as possible, best with infinite dimension, while still allows tractable solution. 
%
%Critically, we show that derivative in \eqref{equ:f00} yields a closed form representation:
%$$  \frac{d}{d\epsilon} \KL(q_{[\epsilon\ff]} ~|| ~ p) \big |_{\epsilon = 0}  = \E_{x\sim q} [\trace(\stein_p \ff(x))], $$Critically, we show that $\F$ to be the unit ball of a reproducing kernel Hilbert space (RKHS) $\H$
%
%
%%$\KL(q_{\vx^t} ~||~ p)$ between the distribution between the distribution $q_{\vx}$ of particles $\vx^t$ and the target distribution decreases with the fastest speed. 
%We take $\F$ to be the unit ball of a reproducing kernel Hilbert space (RKHS) $\H$ associated with positive definite kernel $k(x,x')$. 
%Thanks to an important connection with Stein's identity and kernelized Stein discrepancy, we show that the optimal $\phi$ is given by 
%$\ff(x)$ in RKHS $\H$ associated with positive definite kernel $k(x,x')$, in which case the optimal choice of $\ff(\vx)$ is 

Therefore, the optimization in \eqref{equ:ff00} reduces to 
\begin{align}\label{equ:ksd}
\S(q ~||~ p) \overset{def}{=} \max_{\ff \in \H^d} \{ \E_{x\sim q} [\sumstein_p \ff(x)]  ~~~s.t.~~~~ ||\ff ||_{\H^d} \leq 1\}, 
\end{align}
where $\S(q ~||~ p)$ is the kernelized Stein discrepancy defined in \citet{liu2016kernelized}, which equals zero if and only if $p = q$ under mild regularity conditions. 
Importantly, the optimal solution of \eqref{equ:ksd} yields a closed form% for the optimal solution: of \eqref{equ:ff00}:
 %Further,  it turns out that optimizing \eqref{equ:klstein00} in the unit ball of $\H$ coincides with the variational form of the kernelized Stein discrepancy in \citet{liu2016kernelized},  giving a closed form for the optimal solution of \eqref{equ:ff00}: 
$$
\ff^*(x') \propto  \E_{x\sim q}[\nabla_x \log p(x)k(x,x') + \nabla_x k(x,x')]. 
$$
%This $\ff^*$ can be treated as a gradient of the KL divergence. 
By approximating the expectation under $q$ with the empirical average of the current particles $
\{x_i\}_{i=1}^n$,  SVGD admits a simple form of update: 
\begin{align}\label{equ:update11}
%\begin{split}
%x_i ~ \gets ~ x_i  ~  + ~ \epsilon  \hat\ff^*(x_i) ~~~~~\text{where}~~~~~~ \hat \ff^*(x_i) = \hat \E_{x\in \{x_i\}} [ k(x_i, ~ x) \nabla_{x} \log p(x)  + \nabla_{x} k(x_i, x) ], 
&& &~~~~~~~~~~~ x_i ~ \gets ~ x_i  ~  + ~ \epsilon \Delta x_i, 
~~~~~~~~\forall i = 1, \ldots, n,  \notag
\\
~
&&& \text{where~~~~~}\Delta x_i = \hat \E_{x\in \{x_i\}_{i=1}^n} [  \nabla_{x} \log p(x) k(x, x_i) + \nabla_{x} k(x, x_i) ], 
%x_i ~ \gets ~ x_i  ~  + ~ \epsilon \frac{1}{n} \sum_{x\in \{x_i\}} [ k(x_i, ~ x_j) \nabla_{x} \log p(x)  + \nabla_{x} k(x_i, x) ] . 
%\end{split}
\end{align}
and $\hat\E_{x\sim \{x_i\}_{i=1}^n}[f(x)] = \sum_i f(x_i)/n$. 
The two terms in $\Delta x_i$ play two different roles: 
the term with the gradient $\nabla_x \log p(x)$ drives the particles toward the high probability regions of $p(x)$, 
while the term with $\nabla_x k(x,x_i)$ serves as a repulsive force to encourage diversity;
to see this, consider a stationary kernel $k(x,x') = k(x-x')$, then the second term reduces to $\hat \E_x \nabla_{x} k(x,x_i) = - \hat \E_x \nabla_{x_i} k(x,x_i)$, 
which can be treated as the negative gradient for minimizing the average similarity $\hat \E_x k(x,x_i)$ in terms of $x_i$. 
Overall, this particle update produces diverse points for distributional approximation and uncertainty assessment, and 
also has an interesting ``momentum'' effect in which the particles move collaboratively to escape the local optima.
%See \citet{liu2016stein} for more details. 

It is easy to see from \eqref{equ:update11} that $\Delta x_i$ reduces to the typical gradient $\nabla_x \log p(x_i)$ when there is only a single particle ($n=1$) and $\nabla_x k(x,x_i)$ when $x=x_i$,  
in which case SVGD reduces to the standard gradient ascent for maximizing $\log p(x)$ (i.e., maximum \emph{a posteriori} (MAP)). 

%This update evolves the particles collectively to match the target distribution with the two terms of $\Delta(x_i)$ playing two different roles:    the first term in $\Delta(x_i)$ is theweighted sum of the particles' gradient of the log-density, which drives the particles towards the high probability regions in a collaborative fashion; the second term can be shown to serve as a ``repulsive force'' that makes the particles repel each other to maintain a degree of diversity.
% maintaining a degree of diversity via the second term $\nabla_x k(x_i, x)$ that makes the particles repel each other.

\begin{algorithm}[t]                      % enter the algorithm environment
\caption{Amortized SVGD for Problem~\ref{pro:prob1}}% for Wild Variational Inference}          % give the algorithm a caption
\label{alg:alg1}                           % and a label for \ref{} commands later in the document
\begin{algorithmic}                    % enter the algorithmic environment
\STATE Set batch size $m$, step-size scheme $\{\epsilon_t\}$ and kernel $k(x,x')$. Initialize $\eta^0$. 
\FOR {iteration $t$}
\STATE  Draw random $\{\xi_i\}_{i=1}^m$, calculate $x_i = f(\eta^t;~\xi_i)$, 
and the Stein variational gradient $\Delta x_i$ in \eqref{equ:update11}. 
%\begin{align*}%\label{equ:dxi}
%\Delta x_i = \hat\E_{x\sim\{x_i\}_{i=1}^m} [\log p(x) k(x,x_i) + \nabla_x k(x, x_i)]. 
%\end{align*}
\STATE  Update parameter $\eta$ using \eqref{equ:follow1}, \eqref{equ:follow2} or \eqref{equ:follow3}. 
%$$\eta^{t+1} \gets \eta^{t} + \epsilon ~ \sum_{i=1}^n \partial_\eta f(\eta^t;~\xi_i)  \Delta x_i.$$ 
\ENDFOR
\end{algorithmic}
\end{algorithm}

\section{Amortized SVGD: Towards an Automatic Neural Sampler}
\label{sec:amortizedsvgd}
%\section{Our Method}

%SVGD and other particle-based methods require a large memory to store the particles when  using a large number particles. In addition, 
%Particle methods like SVGD can provide simple and consistent approximation for individual target distributions, 
SVGD and other particle-based methods become inefficient when we need to repeatedly infer
a large number different target distributions for multiple tasks, including online learning or inner loops of other algorithms, 
because they can not improve based on the experience from the past tasks, and may require a large memory to restore a large number of particles. 
We propose to ``amortize SVGD'' by training a neural network $f(\eta;~\xi)$ to mimic the SVGD dynamics, yielding a solution for Problem~\ref{pro:prob1}. 

%We propose to use a neural network $f_{\eta}$ so that the output $x = f(\eta;~\xi)$ with $\xi \sim q_0$ mimics the SVGD dynamics. 
One straightforward way to achieve this is to run SVGD to convergence and 
train $f(\eta;~\xi)$ to fit the SVGD results. 
This, however, requires to run many epochs of fully converged SVGD and can be slow in practice. 
We instead propose an \emph{incremental approach} in which $\eta$ is iteratively adjusted 
so that the network outputs $x = f(\eta;~\xi)$ changes along the Stein variational gradient direction in \eqref{equ:update11} 
in order to decrease the KL divergence between the target and approximation distribution. 
%to follow the change $\Delta x_i$ given by SVGD. 
% to mimic the SVGD updates which can summary 
%the experience from the past tasks concisely, allowing ``learning to draw samples''. 
%In this 
%Many practice applications involve a large number of target distributions with similar structures for which 
%it is inefficient to repeatedly applying SVGD or other tradition inference methods on the individual distributions separately. 
%and it would be highly desirable to intelligent systems that \emph{train itself to make better Bayesian inference over time, and also generalize well to new tasks with similar structures}. 
%becomes inefficient when we have a large number of different target distributions that requires us to run the particle algorithm repeatedly. 
%The different target
%This happens when we need to reason with different observed data or priors, or need to make iterative, real-time inference for online tasks or inner loops of learning algorithms. 
%Therefore, it is highly desirable to develop more intelligent Bayesian inference systems that can adaptively improve its own performance over time, and also generalize well to new tasks with similar structures. 

%Our idea is simple: 
%we want to iteratively adjust $\eta$ such that the black-box output $x = f(\eta;~D, \xi)$ moves along the Stein variational gradient direction $\Delta(x_i)$ in \eqref{equ:update11} in order to decrease the KL divergence between the target and approximation distribution. 
To be specific, denote by $\eta^t$ the estimated parameter at the $t$-th iteration of our method; 
each iteration of our method 
draws a batch of random inputs $\{\xi_i\}_{i=1}^m$ 
and calculate their corresponding output $x_i = f(\eta;~\xi_i)$ based on $\eta^t$; here $m$ is a mini-batch size (e.g., $m=100$). 
%as well as $x'_i  = x_i \epsilon \Delta x_i$%B
The Stein variational gradient $\Delta x_i$ in 
\eqref{equ:update11} would then ensure that $x'_i = x_i + \epsilon \Delta x_i$ forms a better approximation of the target distribution $p$. 
Therefore, we should adjust $\eta$ to make its output matches $\{x'_i\}$, 
that is, we want to update $\eta$ by
\begin{align}\label{equ:follow1}
\eta^{t+1} \gets  \argmin_\eta  \sum_{i=1}^m ||  f(\eta;~\xi_i)  - x_i' ||_2^2,  ~~~~~~ \text{where} ~~~~~~  x_i'  = x_i  + \epsilon \Delta x_i. 
\end{align}
%. 
See Algorithm~\ref{alg:alg1} for the summary of this procedure. If we assume $\epsilon$ is very small, then 
\eqref{equ:follow1} reduces to a least square optimization. To see this, note that 
$f(\eta;~\xi_i) \approx f(\eta^t;~\xi_i) + \partial_\eta f(\eta^t;~\xi_i) (\eta - \eta^t)$ by Taylor expansion. 
Since $x_i = f(\eta^t;~\xi_i)$, we have
%Therefore, 
$$
 ||  f(\eta;~\xi_i)  - x_i' ||_2^2 \approx || \partial_\eta f(\eta^t;~\xi_i) (\eta  - \eta^t)  - \epsilon \Delta x_i  ||_2^2. 
$$
%This allows us to approximate the update i
As a result, \eqref{equ:follow1} reduces to the following least square optimization: 
\begin{align}\label{equ:follow2}
\eta^{t+1} \gets \eta^t + \epsilon \Delta \eta^t, 
\text{~~~~~where~~~~~}
\Delta \eta^t = \argmin_{\delta}  \sum_{i=1}^m   || \partial_\eta f(\eta^t;~\xi_i)  \delta   -  \Delta x_i  ||_2^2. 
\end{align}
%Although $\Delta \eta^t$ can be solved with a least square optimization, 
Update \eqref{equ:follow2} can still be computationally expensive because of the matrix inversion. 
% in each iteration. 
We can derive a further approximation by performing only one step of gradient descent of \eqref{equ:follow1} (or \eqref{equ:follow2}), which gives 
\begin{align}\label{equ:follow3}
\eta^{t+1} \gets \eta^t + \epsilon  \sum_{i=1}^m   \partial_\eta f(\eta^t;~\xi_i) \Delta x_i. 
\end{align}
%based on update \eqref{equ:follow3}. 
%We find this update works efficiently in practice. It is shown in Algorithm~\ref{alg:alg1}

%In this way, everything is nice

Although update \eqref{equ:follow3} is derived as an approximation of \eqref{equ:follow1}-\eqref{equ:follow2}, 
it is computationally faster and we find it works very effectively in practice; 
this is because when $\epsilon$ is small, 
one step of gradient update can be sufficiently close to the optimum.  

%Update \eqref{equ:follow3} has a particularly 
%Intuitively speaking, 
Update \eqref{equ:follow3} also has a simple and intuitive form: \eqref{equ:follow3} can be thought as \emph{a ``chain rule'' that back-propagates the Stein variational gradient to the network parameter $\eta$}. 
%making the random sample generated by $z = f(\eta;~\xi)$ mimic the behavior of SVGD and hence move towards the target distribution. 
%Although $\Delta x_i$ is not a gradient in the typical sense, it can be interpreted as a form of functional gradient with which the update of $\eta$ can be justified theoretically. 
This can be justified by considering the special case when we use only a single particle $(n=1)$ 
in which case $\Delta x_i$ in \eqref{equ:update11} reduces to the typical gradient $\nabla_x \log p(x_i)$ of $\log p(x)$, 
%In particular, note that if we just use a single particle for each $D$, our method reduces 
and update \eqref{equ:follow3} reduces to the typical gradient ascent for maximizing 
%a typical stochastic gradient ascent for maximizing 
$$\E_{\xi}[\log p(f(\eta;~\xi))],$$  
in which case $f(\eta;~\xi)$ is trained to maximize $\log p(x)$ (that is, \emph{learning to optimize}), 
instead of \emph{learning to draw samples from $p$} for which it is crucial to use Stein variational gradient $\Delta x_i$ to diversify the network outputs. 
%instead of drawing sampling from $p$ which requires to diversify the network output. 
%the method by \citet{andrychowicz2016learning} for learning to optimize using gradient descent.

Update \eqref{equ:follow3} also has a close connection with the typical variational inference with the reparameterization trick \citep{kingma2013auto}. Let $q_\eta(x)$ be the density function of $x = f(\eta;~\xi)$, $\xi\sim q_0$. Using the reparameterization trick, the gradient of $\KL(q_\eta~||~p)$ w.r.t. $\eta$ can be shown to be 
$$
\nabla_\eta\KL(q_{\eta}~||~p) = -\E_{\xi \sim q_0}[\partial_\eta f(\eta;~\xi)(\nabla_x \log p(x) - \nabla_x \log q_\eta(x))]. 
$$
With $\{\xi_i\}$ i.i.d. drawn from $q_0$ and $x_i = f(\eta;~\xi_i), ~\forall i$, the standard stochastic gradient descent for minimizing the KL divergence is 
%can be performed by 
\begin{align}\label{equ:rep}
\eta^{t+1} \gets \eta^t +  \sum_i \partial_\eta f(\eta^t;~\xi_i) \tilde \Delta x_i, ~~~~ \text{where} ~~~~ 
\tilde \Delta x_i = \nabla_x \log p(x_i) - \nabla_x \log q_\eta(x_i). 
\end{align}
This is similar with \eqref{equ:follow3}, but replaces the Stein gradient $\Delta x_i$ defined in \eqref{equ:update11} with 
$\tilde \Delta x_i$. The advantage of using $\Delta x_i$ is that it does not require to explicitly calculate $q_\eta$,
and hence admits a solution to Problem 1 in which $q_\eta$ is not computable for complex network $f(\eta; ~\xi)$ and unknown input distribution $q_0$. 
Further insights can be obtained by noting that 
\begin{align}
\label{equ:tmp}
\Delta x_i 
& \approx \E_{x\sim q}[\nabla_x \log p(x)k(x,x_i) + \nabla_xk(x,x_i)] \notag \\
& =  \E_{x\sim q}[(\nabla_x \log p(x) - \nabla_x \log q(x))k(x,x_i)]  \\
& = \E_{x\sim q} [( \tilde \Delta x) k(x, x_i)],  \notag
%& = \E_{x\sim q} [k(x, x_i) \tilde \Delta x]  \notag
%\\&\approx  \sum_j k(x_j, x_i) \tilde \Delta x_i \notag
\end{align}
where \eqref{equ:tmp} is obtained by using Stein's identity \eqref{equ:steinid}. 
Therefore, %$\Delta x_i$ is approximately $\Delta x_i$ multiplied by a positive definite matrix $[k(x_i, x_j)]_{ij}$ and hence with positive inner product.% $\Delta x_i, \tilde \Delta x_i\la $. 
$\Delta x_i$ can be treated as a kernel smoothed version of $\tilde \Delta x_i $. 
%It is also possible to get $q_\eta(x)$-free (wild) variational inference by directly approximating $\nabla_x\log q_\eta(x)$ based on 

\section{Amortized MLE for Generative Adversarial Training}
Our method allows us to design efficient approximate sampling methods 
adaptively and automatically, and enables a host of novel applications. 
In this paper, we apply it in an amortized MLE method for training deep generative models.   
%(2) automatic hyper-parameter tuning for Bayesian inference. 
%including optimizing hyperparameters in traditional inference algorithms, 
%speeding up repeated inference tasks for online or streaming settings, or inner loops of the algorithms (such as maximum likelihood learning). 
%By leveraging state-of-the-art network architectures such as RNN and ResNet and fully optimizing the  
%or eventually replacing hand-designed infernece methods with more effiicent ones that trained on past tasks and improve adaptively over time. 

\begin{figure}[h]
\centering
\includegraphics[width=0.9\textwidth]{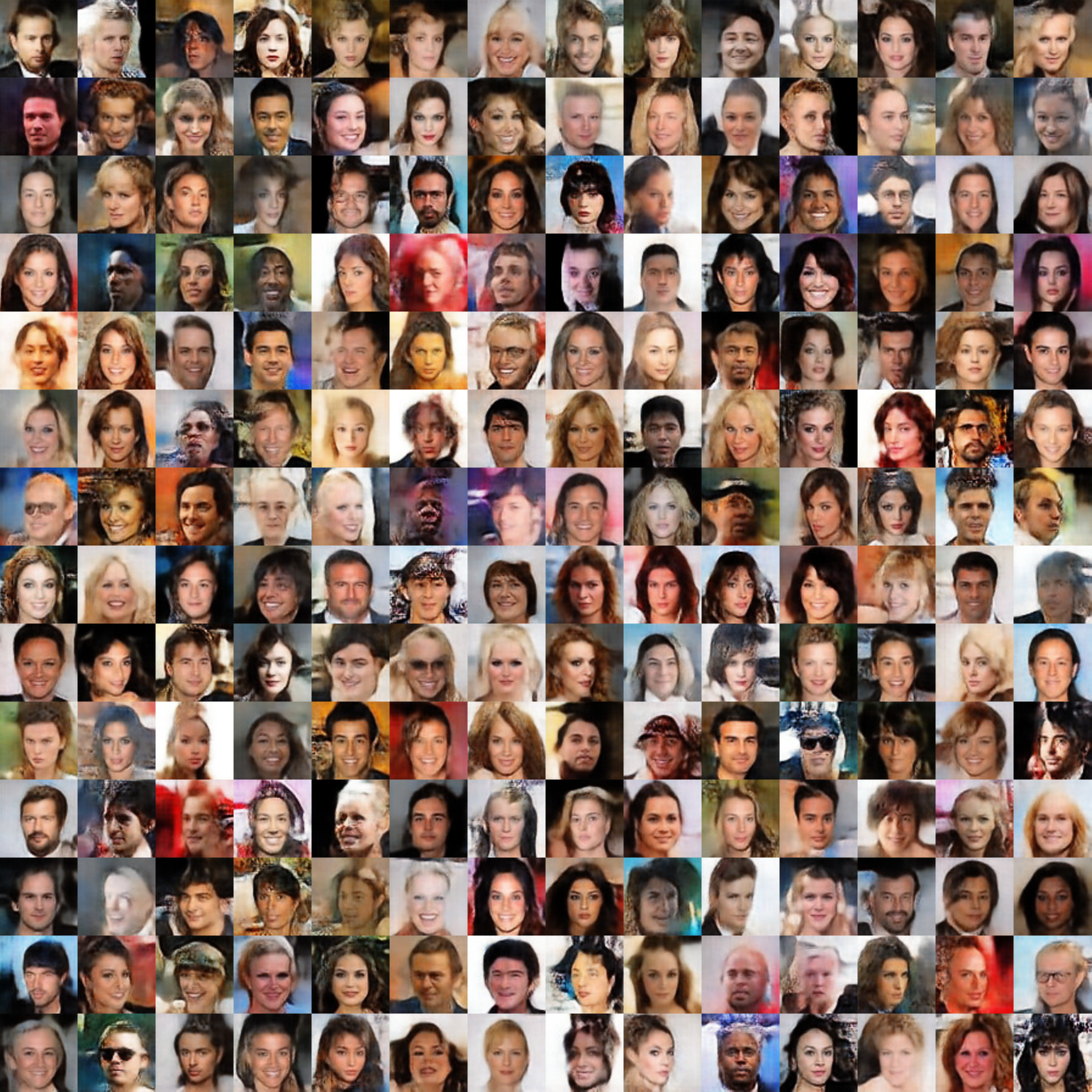}  
\caption{More images generated by SteinGAN on CelebA.}
\label{fig:facemore}
\end{figure}

%\emph{Amortized maximum likelihood estimation (MLE) for Deep Generative Models} %Generative Adversarial Networks based
%\subsection{Amortized MLE for Generative Adversarial Training} %Generative Adversarial Networks based
 Maximum likelihood estimator (MLE) provides a fundamental approach for learning probabilistic models from data, 
but can be computationally prohibitive on distributions for which drawing samples or computing likelihood is intractable due to the normalization constant. 
Traditional methods such as MCMC-MLE use hand-designed methods (e.g., MCMC) to approximate the intractable likelihood function but do not work efficiently in practice. 
We propose to adaptively train a generative neural network to draw samples from the distribution during MLE training, which not only provides computational advantage, and also allows us to generate realistic-looking images competitive with, or better than the state-of-the-art generative adversarial networks (GAN) \citep{goodfellow2014generative, radford2015unsupervised} (see Figure~\ref{fig:mnist}-\ref{fig:facemore}).  
%See our empirical results in Figure~\ref{fig:face}.
% computationally intractable normalization constant. %(or its derivative). 

To be specific, denote by $\{x_{i,obs}\}$ a set of observed data. 
We consider the maximum likelihood training of energy-based models of form 
$$
p(x|\theta) = \exp(-\phi(x, \theta) - \Phi(\theta)), ~~~~~ \Phi(\theta) = \log \int \exp(-\phi(x,\theta))dx,
$$
where $\phi(x; ~\theta)$ is an energy function for $x$ indexed by parameter $\theta$ and $\Phi(\theta)$ is the log-normalization constant. %partition function. 
The log-likelihood function of $\theta$ is %based on maximizing the log likelihood function, 
$$
L(\theta)  =\frac{1}{n}\sum_{i=1}^n\log p(x_{i,obs} | \theta),
$$
whose gradient is %ascent update is 
\begin{align*}%
\nabla_\theta L(\theta) = -  \hat\E_{obs} [\partial_\theta \phi (x; \theta) ] +    \E_\theta [\partial_\theta \phi(x; \theta) ], 
%\nabla_\theta L(\theta) =  \la\nabla_\theta f(x; \theta) \ra_{Data} -   \la  \nabla_\theta f(x; \theta) \ra_{\theta}, 
\end{align*}
where $\hat\E_{obs}[\cdot]$ and $\E_\theta[\cdot]$ denote the empirical average on the observed data $\{x_{i,obs}\}$
and the expectation under model $p(x |\theta)$, respectively.  
The key computational difficulty is to approximate the model expectation $\E_{\theta}[\cdot]$. 
To address this problem, 
we use a generative neural network $x = f(\eta;~\xi)$ trained by Algorithm~\ref{alg:alg1} 
to approximately sample from $p(x|\theta)$, yielding a gradient update for $\theta$ of form
\begin{align}\label{equ:updatetheta}
\theta \gets \theta + \epsilon \hat\nabla_\theta L(\theta),  &\text{}& 
 \hat \nabla_\theta L(\theta) =  - \hat\E_{obs} [\partial_\theta \phi (x; \theta) ] +   \hat\E_{\eta} [\partial_\theta \phi(x; \theta) ], 
%\nabla_\theta L(\theta) =  \la\nabla_\theta f(x; \theta) \ra_{Data} -   \la  \nabla_\theta f(x; \theta) \ra_{\theta}, 
\end{align}
where $\hat \E_{\eta}$ denotes the empirical average on $\{x_i\}$ where $x_i = f(\eta;~\xi_i)$, 
$\{\xi_i\}\sim q_0$. 
As $\theta$ is updated by gradient ascent, 
$\eta$ is successively updated via Algorithm~\ref{alg:alg1} to \emph{follow} $p(x|\theta)$. 
See Algorithm \ref{alg:gan}. 

We call our method \emph{SteinGAN}, because it can be intuitively
%Our process can be 
interpreted as an adversarial game between 
the generative network $f(\eta;~\xi)$ and 
the energy model $p(x|\theta)$ which serves as a discriminator:
The MLE gradient update of $p(x|\theta)$ effectively decreases the energy of the training data and increases the energy of the simulated data from $f(\eta;~\xi)$, while the SVGD update of $f(\eta;~\xi)$ decreases the energy of the simulated data to fit better with $p(x|\theta)$.  
%Meanwhile, our procedure is still a principled maximum likelihood procedure and can be more stable than the original GAN \citep{goodfellow2014generative} that attends to find a Nash equilibrium. 
Compared with the traditional methods based on MCMC-MLE or contrastive divergence, we \emph{amortize the sampler as we train}, which gives much faster speed and simultaneously provides a high quality generative neural network that can generate realistic-looking images; see \citet{kim2016deep} for a similar idea and discussions. 
%We find that 
%Traditional approaches \citep{ngiam2011learning, xie2016theory} for training energy-based models are often based on variants of MCMC-MLE or contrastive divergence \citep{geyer1991markov, hinton2002training, tieleman2008training} can not generate realistic-looking images. 
%We will perform comprehensive tests on various applications, including image generation and semi-supervised learning. 

\begin{algorithm}[t]                      % enter the algorithm environment
\caption{Amortized MLE as Generative Adversarial Learning}% for Wild Variational Inference}          % give the algorithm a caption
\label{alg:gan}                           % and a label for \ref{} commands later in the document
\begin{algorithmic}                    % enter the algorithmic environment
\STATE {\bf Goal:} MLE training for energy model $p(x|\theta) = \exp(-\phi(x,\theta) - \Phi(\theta))$.
%the regularized likelihood $\sum_i \log \exp(-\phi(x, \theta)-\Phi(\theta)) + R(\theta)$. 
\STATE Initialize $\eta$ and $\theta$. 
\FOR {iteration $t$}
%\FOR {iteration $s$ (inner loop for updating $\eta$)}
\STATE {\bf Updating $\eta$:} Draw $\xi_i\sim q_0$, $x_i = f(\eta;~\xi_i)$; update $\eta$ using \eqref{equ:follow1}, \eqref{equ:follow2} or \eqref{equ:follow3} with $p(x)=p(x|\theta)$. Repeat several times when needed. 
%\ENDFOR
\STATE 
 {\bf Updating $\theta$:}  Draw a mini-batch of observed data $\{x_{i,obs}\}$, and simulated data $x_i = f(\eta;~\xi_i)$, update $\theta$ by \eqref{equ:updatetheta}. 
%\begin{align}\label{equ:updatetheta}
%\theta \gets \theta - \hat\E_{obs}[ \nabla_\theta \phi (x,\theta) ] + 
%\hat\E_{\eta} [\nabla_\theta \phi(x, \theta)]. % + \nabla R(\theta).  
%\end{align}
\ENDFOR
\end{algorithmic}
\end{algorithm}

%Figure ~\ref{} - \ref{} show the result of our SteinGAN. 
\section{Empirical Results}  \label{sec:gan}

\begin{figure}[t]
\centering
\begin{tabular}{ccc}
\raisebox{8.3em}{
\renewcommand{\arraystretch}{1.5}
\newcommand{\tmpfnt}{\small}
\hspace{-.04\textwidth}
\begin{tabular}{r}
\tmpfnt airplane \\
\tmpfnt automobile \\
\tmpfnt bird \\
\tmpfnt cat \\
\tmpfnt deer \\
\tmpfnt dog \\
\tmpfnt frog \\
\tmpfnt horse \\
\tmpfnt ship \\
\tmpfnt truck
\end{tabular}} & 
\hspace{-.04\textwidth}\includegraphics[width=.42\textwidth]{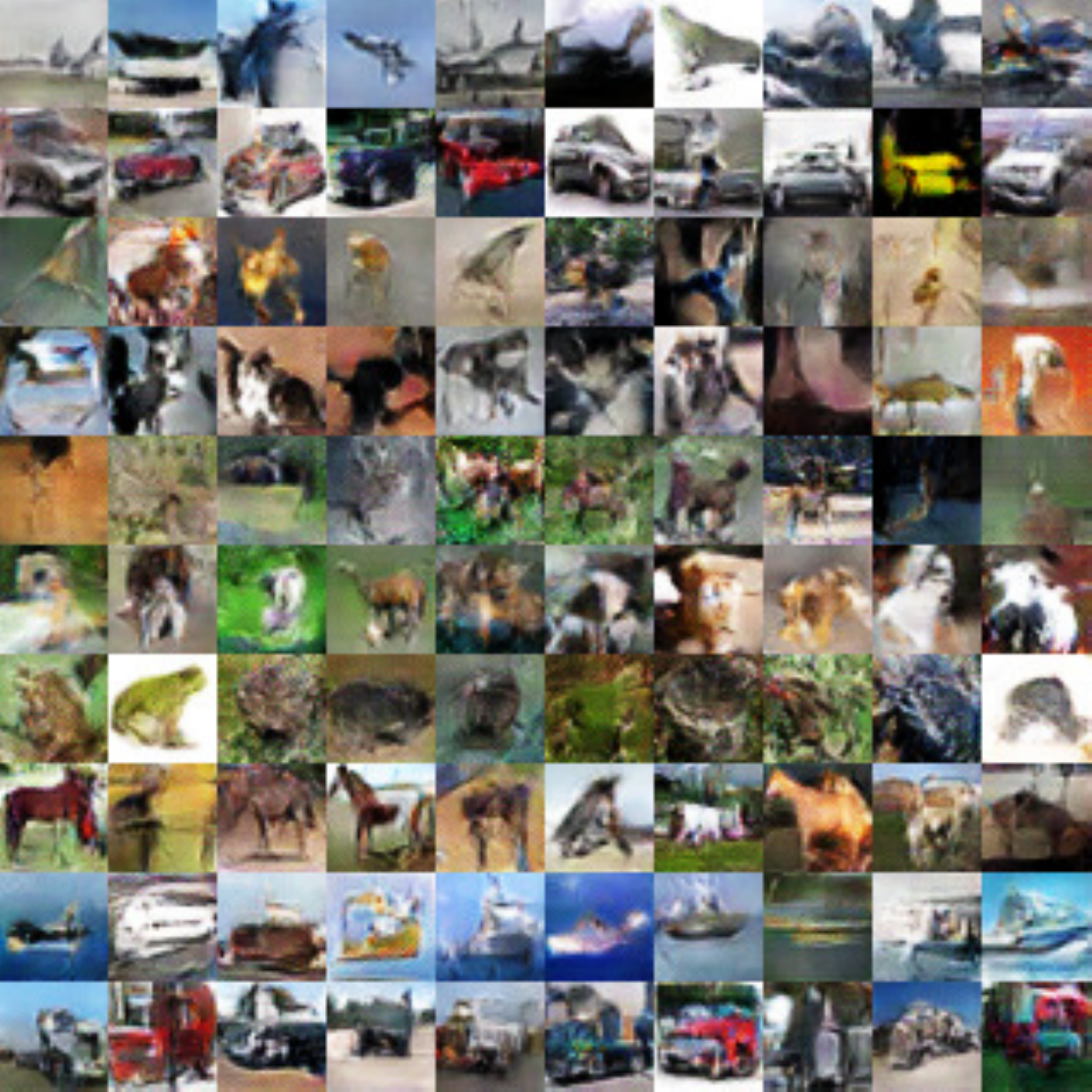}  & 
\includegraphics[width=.42\textwidth]{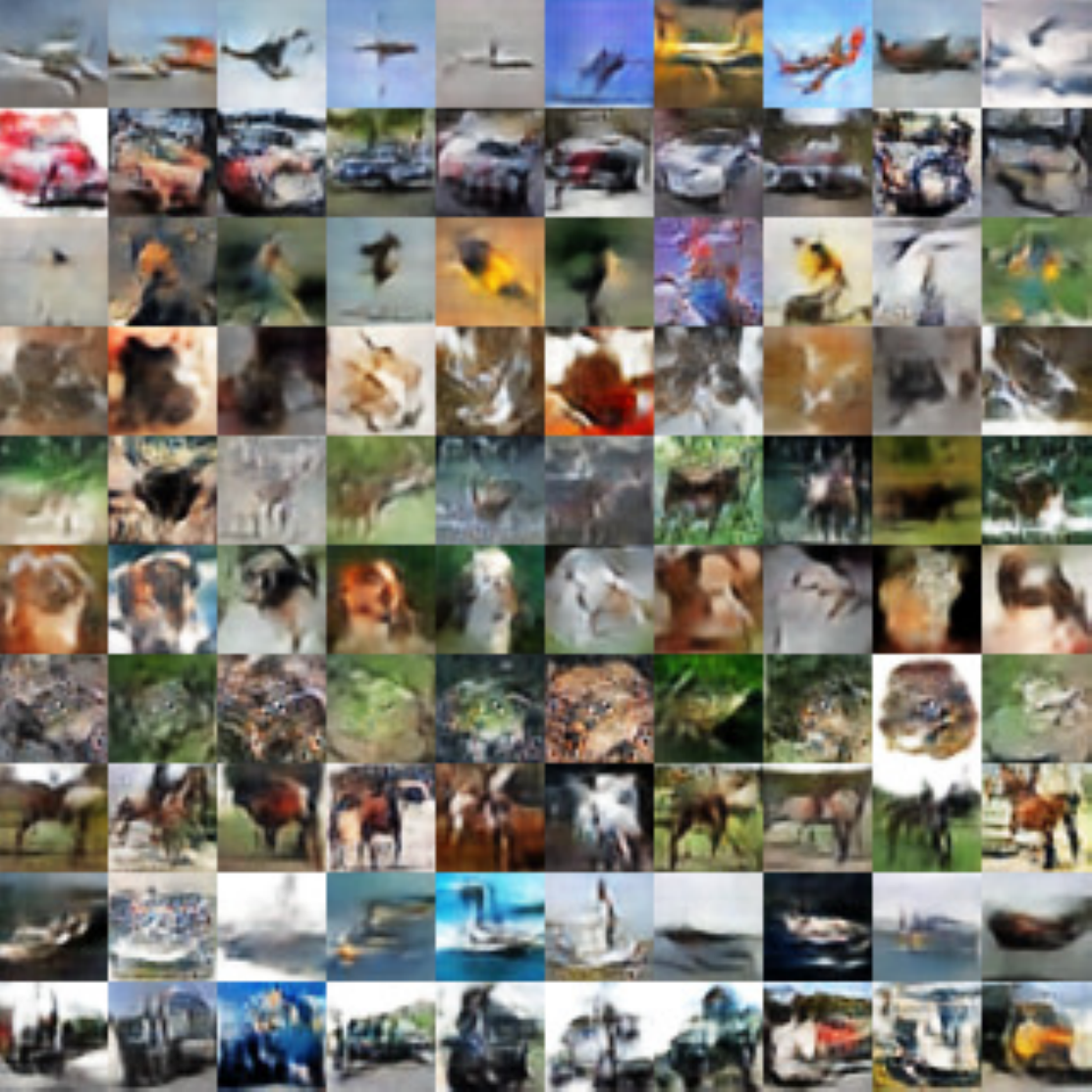}  \\
& DCGAN 
%& SteinGAN (w/o kernel) 
& SteinGAN \\
\vspace{5\baselineskip}
\end{tabular}
\renewcommand{\arraystretch}{1.2}
\begin{tabular}{|l|c|c|c|c|}
\hline
\multicolumn{5}{|c|}{Inception Score} \\
\hline
 & Real Training Set & 500 Duplicate  & DCGAN  & SteinGAN \\
\hline
%Model Trained on ImageNet  & 11.237 & 11.100 & 6.581 & 6.209 \\
Model Trained on ImageNet  & 11.237 & 11.100 & 6.581 & 6.351 \\ %6.3514
\hline
%Model Trained on CIFAR-10 & 9.848 & 9.807 & 7.193 & 7.065 \\
Model Trained on CIFAR-10 & 9.848 & 9.807 & 7.368  & 7.428 \\  %7.368 % 7.193
\hline 
\end{tabular}  \\
\renewcommand{\arraystretch}{1.2}
\begin{tabular}{|c|c|c|c|}
\hline
\multicolumn{4}{|c|}{Testing Accuracy} \\
\hline
Real Training Set & 500 Duplicate  & DCGAN  & SteinGAN \\
\hline
92.58 \% &  44.96 \% & 44.78 \%  & 63.81 \%\\  %62.72 %63.81
\hline
\end{tabular}
\caption{Results on CIFAR-10. ``500 Duplicate'' denotes  500 images randomly subsampled from the training set, each duplicated 100 times.
Upper: images simulated by DCGAN and SteinGAN (based on joint model \eqref{equ:pxy}) conditional on each category.  
Middle: inception scores for samples generated by various methods (all with 50,000 images) on inception models trained on ImageNet and CIFAR-10, respectively. 
Lower: testing accuracy on real testing set when using 50,000 simulated images to train ResNets for classification. SteinGAN achieves higher testing accuracy than DCGAN. We set $m=1$ and $\gamma=0.8$.}%even though the images on in the upper panel look almost equally good visually.}
\label{fig:cifar10}
\end{figure}

\begin{figure}[t]
\centering
\begin{tabular}{cc}
\includegraphics[width=0.45\textwidth]{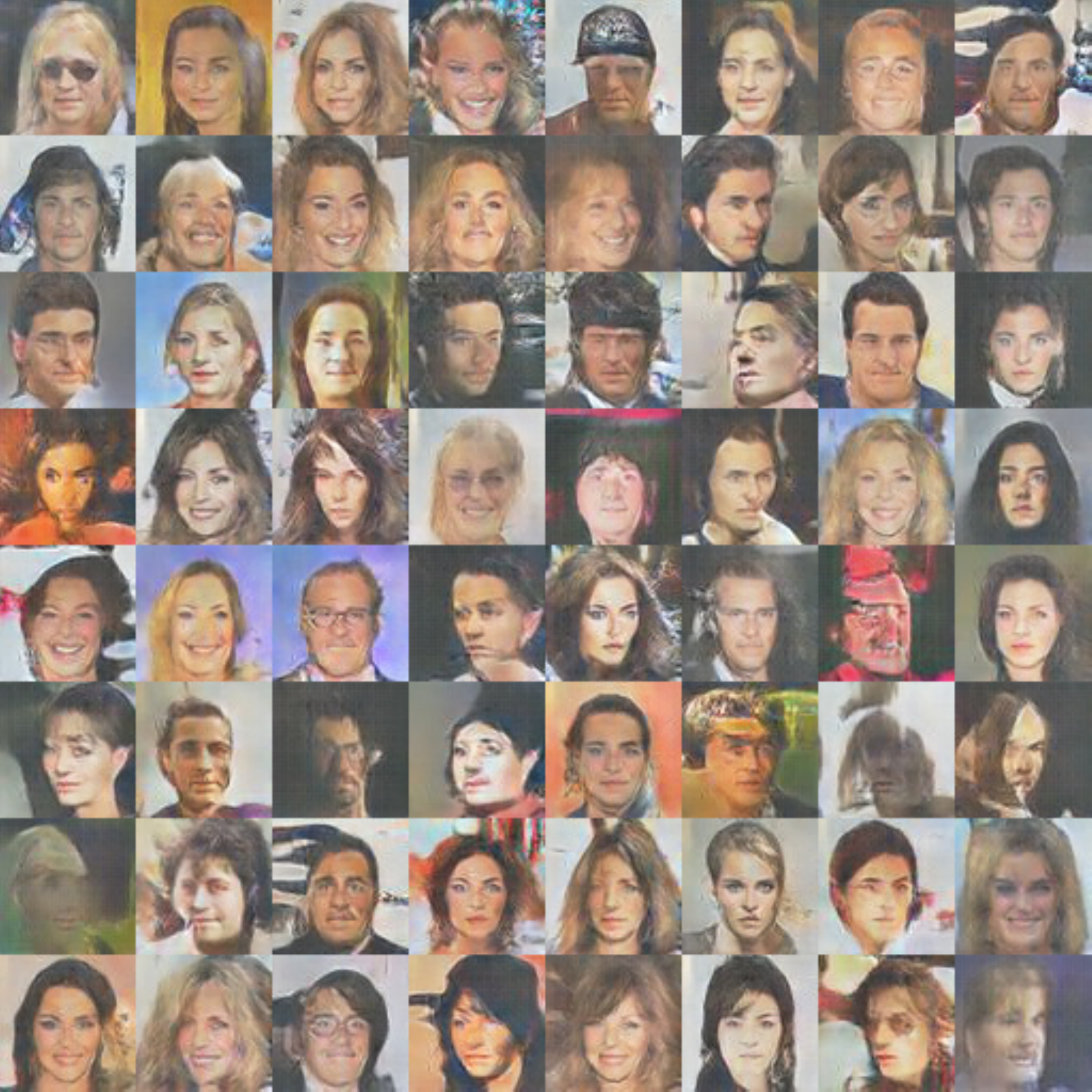} & 
\includegraphics[width=0.45\textwidth]{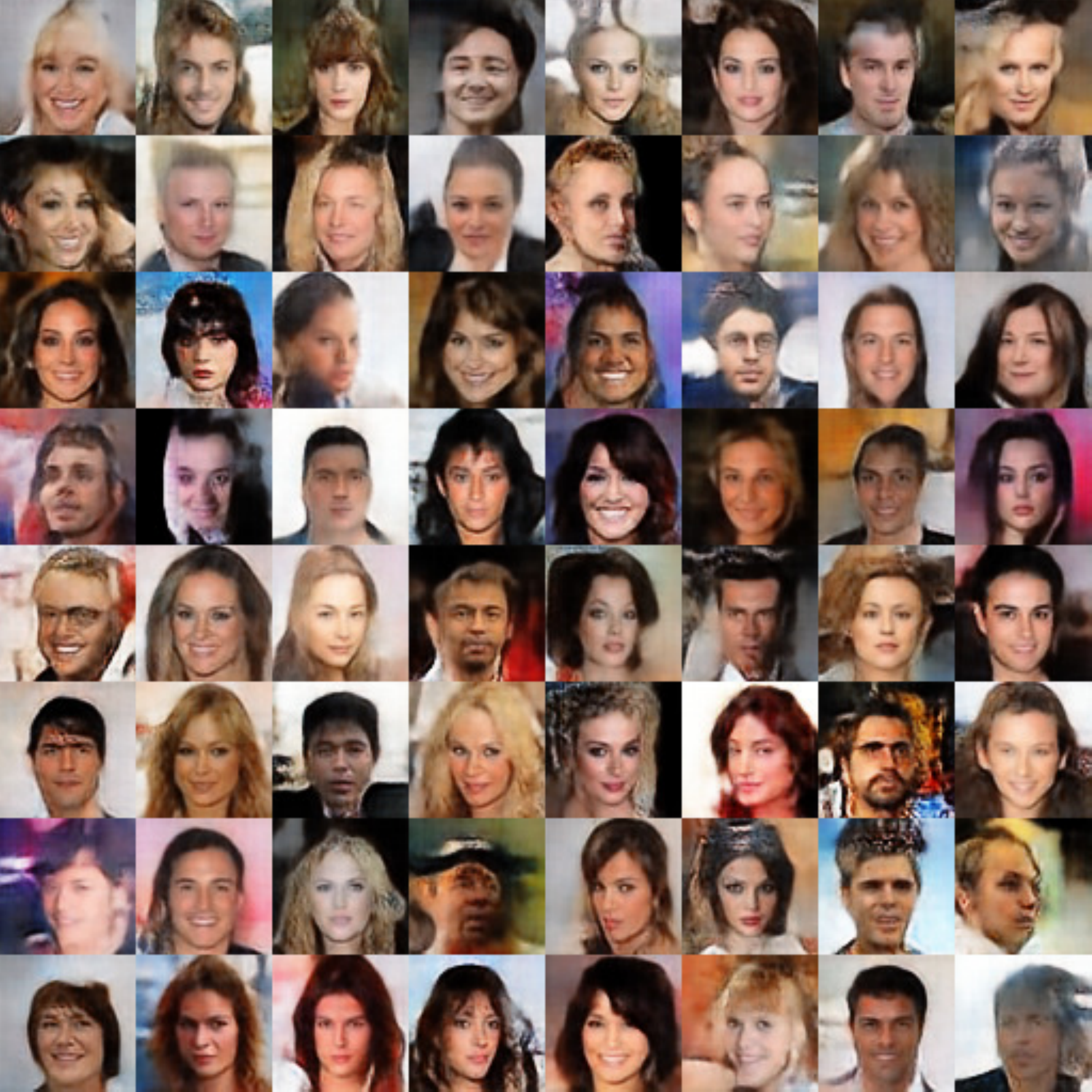} \\
DCGAN & SteinGAN \\
\end{tabular}
\begin{tabular}{c}
\includegraphics[width=0.85\textwidth]{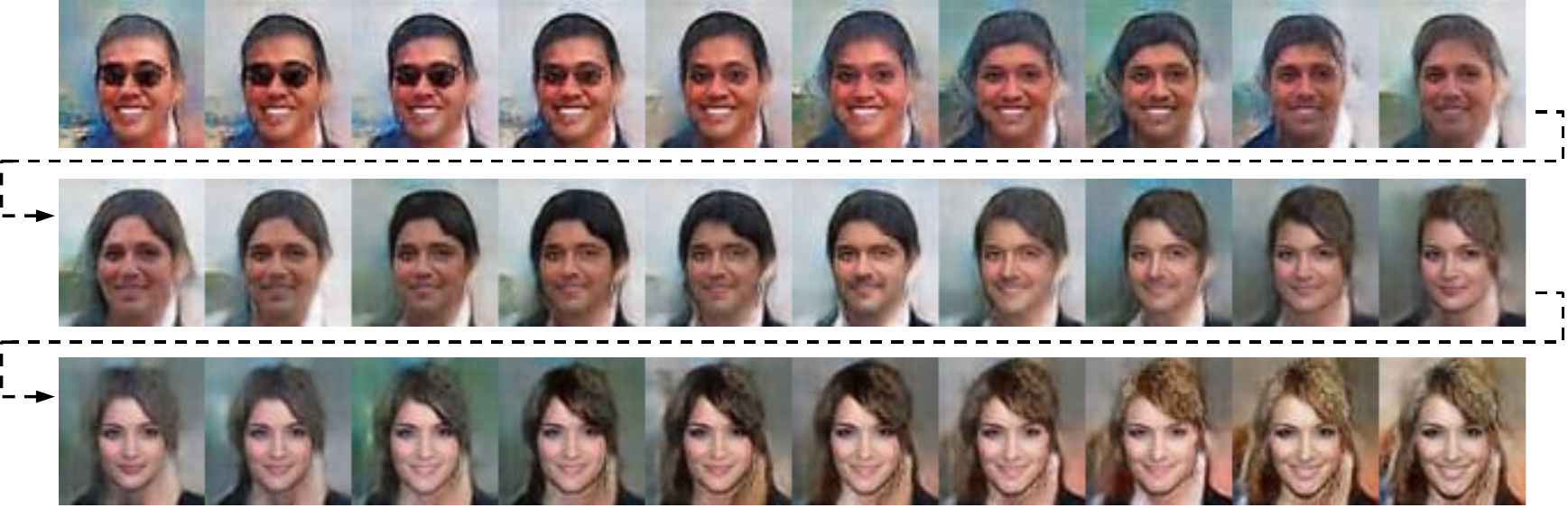} \\
\end{tabular}
\caption{Results on CelebA. Upper: images generated by DCGAN and our SteinGAN. Lower: images generated by SteinGAN when performing a random walk $\xi\gets \xi + 0.01\times\mathrm{Uniform}([-1,1])$ on the random input $\xi$; we can see that a man with glasses and black hair gradually changes to a woman with blonde hair. 
See Figure~\ref{fig:facemore} for more examples. }
%It can be observed that a generation of a man with glasses is gradually changed to a woman with blonde hair, all of which make a lot of sense.}
\label{fig:face}
\end{figure}

We evaluated our SteinGAN on four datasets,  
MNIST, CIFAR-10, CelebA \citep{liu2015faceattributes}, and Large-scale Scene Understanding (LSUN) \citep{yu2015lsun}, on which we find our method tends to generate realistic-looking images competitive with, sometimes better than DCGAN \citep{radford2015unsupervised} (see Figure~\ref{fig:cifar10} - Figure~\ref{fig:face}). 
% mnist \footnote{\url{http://yann.lecun.com/exdb/mnist/}}
% cifar10 \footnote{\url{https://www.cs.toronto.edu/~kriz/cifar.html}}
% celeba  \footnote{\url{http://mmlab.ie.cuhk.edu.hk/projects/CelebA.html}}
% lsun \footnote{\url{http://lsun.cs.princeton.edu/2016/}}
%In particular, we find we generate better images than DCGAN on CelebA (Figure~\ref{fig:face}), and our simulated CIFAR-10 images achives better testing classification accuracy when used as a training data (see Figure~\ref{fig:cifar10}). 
%generative adversarial networks. 
%See Appendix~\ref{sec:gan} for more information. 
%We will provide code to reproduce our experiments. 
Our code is available at \url{https://github.com/DartML/SteinGAN}.

%\subsection{Implementation}
\newcommand{\enc}{\mathrm{E}}
\newcommand{\dec}{\mathrm{D}}
\paragraph{Model Setup} In order to generate realistic-looking images, we define our energy model based on an autoencoder: 
\begin{align}\label{equ:px}
%p(x|\theta) \propto \exp(-\phi(x,\theta)),~~~~~ \text{where~~~~}  \phi(x, \theta) = \alpha || x -   dec(enc(x))  |, 
p(x|\theta) \propto \exp(-  || x -   \dec(\enc(x;~ \theta);~\theta) ||), 
\end{align}
where $x$ denotes the image. 
This choice is motivated by Energy-based GAN \citep{zhao2016energy} in which the autoencoder loss is used as a discriminator but without a probabilistic interpretation. 
We assume $f(\eta;~\xi)$ to be a neural network whose input $\xi$ is a $\dilincheck{100}$-dimensional random vector drawn by $\mathrm{Uniform}([-1,1])$. 
The positive definite kernel in SVGD is defined by the RBF kernel on the hidden representation obtained by the autoencoder in \eqref{equ:px}, that is, 
$$
k(x, x') = \exp(-\frac{1}{h^2} ||\enc(x;~\theta) - \enc(x';~\theta)||^2). 
$$
As it is discussed in Section~\ref{sec:amortizedsvgd}, the kernel provides a repulsive force 
to produce an amount of variability required for generating samples from $p(x)$. 
%to encourage diversity on the generated samples. % via the term $\nabla_x k(x,x')$ in \eqref{equ:update11}. 
This is similar to the heuristic repelling regularizer in \citet{zhao2016energy} and the batch normalization based regularizer in \citet{kim2016deep}, but is derived in a more principled way. 
We take the bandwidth to be $h = \dilincheck{0.5}\times \mathrm{med}$, where $\mathrm{med}$ is the median of the pairwise distances between $\enc(x)$ on the image simulated by $f(\eta;~ \xi)$. 
This makes the kernel change adaptively based on both $\theta$ (through $\E(x;~\theta)$) and $\eta$ (through bandwidth $h$). 

%For MNIST and CIFAR-10, each image $x$ also has a discrete label $y$, and we train a joint model on $(x,y)$:
Some datasets include both images $x$ and 
their associated discrete labels $y$. In these cases, we train a joint energy model on $(x,y)$ 
to capture both the inner structure of the images and its predictive relation with the label,  allowing us to simulate images 
with a control on which category it belongs to. Our joint energy model is defined to be 
%conditional on each categorial. 
\begin{align}\label{equ:pxy}
p(x, y|\theta) \propto \exp\big \{-  || x -   \dec(\enc(x;~\theta);~\theta) || - \max[m,~\sigma(y, ~ \enc(x;~\theta))]\big \},  %\red{dilin: square?}
\end{align}
where $\sigma(\cdot,\cdot)$ is the cross entropy loss function of a fully connected output layer. 
In this case, our neural sampler first draws a label $y$ randomly according to the empirical counts in the dataset, 
and then passes $y$ into a neural network together with a $100\times 1$ random vector $\xi$ to generate image $x$. 
This allows us to generate images for particular categories by controlling the value of input $y$. 
%We set $\beta=\dilincheck{?}$ and $m$ by \dilincheck{?}. 
%where $\sigma$ denotes  a fully connected layer.  

\paragraph{Stabilization}
In practice, we find it is useful to modify \eqref{equ:updatetheta} to be 
\begin{align}\label{equ:disc}
\theta \gets \theta - \epsilon \hat\E_{obs}[ \nabla_\theta \phi (x,\theta) ] + 
\epsilon (1-\gamma) \hat\E_{\eta} [\nabla_\theta \phi(x, \theta)]. 
\end{align}
where $\gamma$ is a discount factor (which we take to be $\gamma = 0.7$). 
%This is equivalent to MAP of $\theta$ with a conjugate prior: %
This is equivalent to maximizing a regularized likelihood: 
$$
\max_\theta  \{ \log   p(x |\theta)  +  \gamma \Phi(\theta)\}
$$
where $\Phi(\theta)$ is the log-partition function; note that $\exp( \gamma \Phi(\theta))$ is a conjugate prior of $p(x|\theta)$. 

We initialize the weights of both the generator and discriminator from Gaussian distribution $\mathcal{N}(0,0.02)$, 
and train them using Adam \citep{kingma2014adam} with a learning rate of $0.001$ for the generator and $0.0001$ for the energy model (the discriminator).  
%We used the Adam \citep{kingma2014adam} optimizer to train both models with mini-batch stochastic gradient descent.
%To achieve good performance, it is important to make sure that generator and discriminator are approximately aligned during training. 
In order to keep the generator and discriminator approximately aligned during training, 
we speed up the MLE update \eqref{equ:disc} of the discriminator (by increasing its learning rate to $0.0005$) when the energy of the real data batch is larger than the energy of the simulated images, 
% during adversarial training, 
while slow down it (by freezing the MLE update of $\theta$ in \eqref{equ:disc}) if the magnitude of the energy difference between the real images and the simulated images goes above a threshold of 0.5.
%Additionally, learning is frozen for the discriminator if the magnitude of the energy difference between real images and fake samples goes above a threshold 0.5.
%We also stabilize the algorithm by .xxxx. 
We used the bag of architecture guidelines for stable training suggested in DCGAN \citep{radford2015unsupervised}.

\begin{comment}
We tested our SteinGAN on four datasets,  
MNIST \footnote{\url{http://yann.lecun.com/exdb/mnist/}}, CIFAR-10 \footnote{\url{https://www.cs.toronto.edu/~kriz/cifar.html}}, CelebA \footnote{\url{http://mmlab.ie.cuhk.edu.hk/projects/CelebA.html}}, and Large-scale Scene Understanding (LSUN) \footnote{\url{http://lsun.cs.princeton.edu/2016/}}, on which we find our method tends to generate realistic-looking images competitive with DCGAN \citep{radford2015unsupervised} (see Figure~\ref{fig:face}-Figure~\ref{fig:cifar10}). 
In particular, we find we generate better images than DCGAN on CelebA (Figure~\ref{fig:face}), 
and our simulated CIFAR-10 images achieves better testing classification accuracy when used as a training data (see Figure~\ref{fig:cifar10}). 
%generative adversarial networks. 
See Appendix~\ref{sec:gan} for more information. 
We will provide code to reproduce our experiments.  
\end{comment}

\begin{figure}[htb]
\centering
\begin{tabular}{cc}
\includegraphics[width=0.3\textwidth]{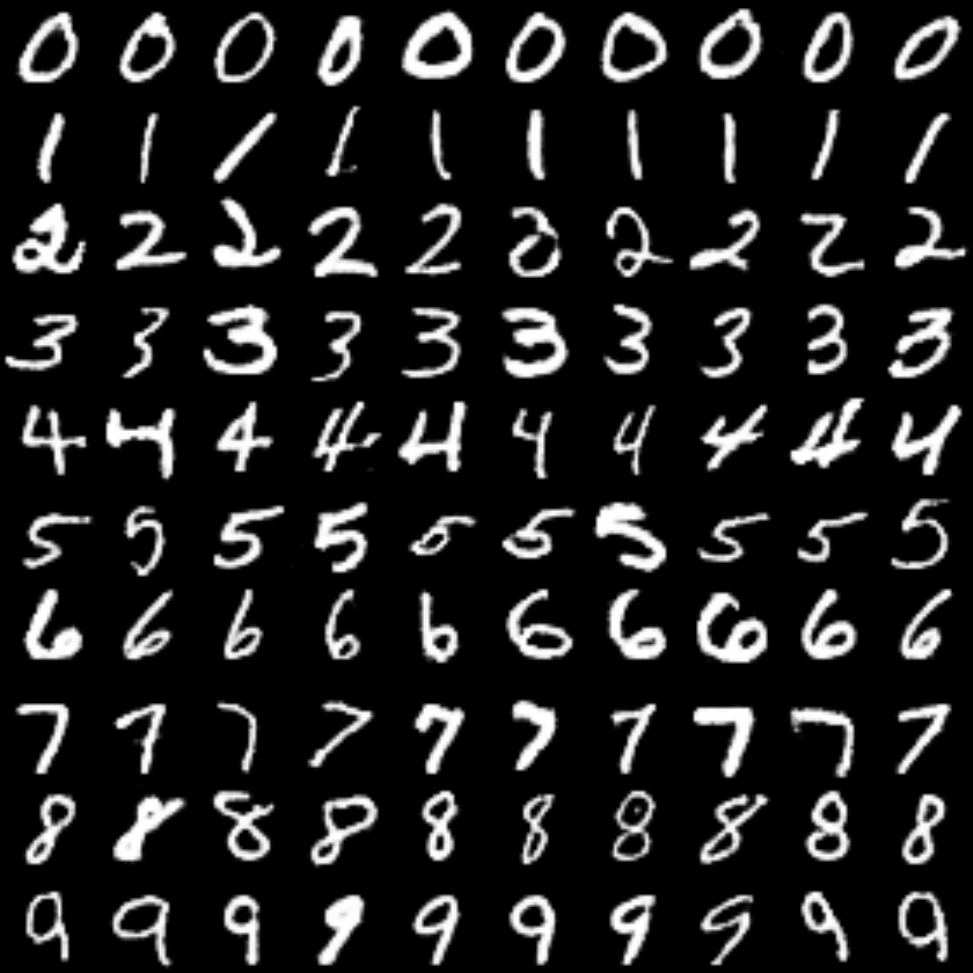} & 
\includegraphics[width=0.3\textwidth]{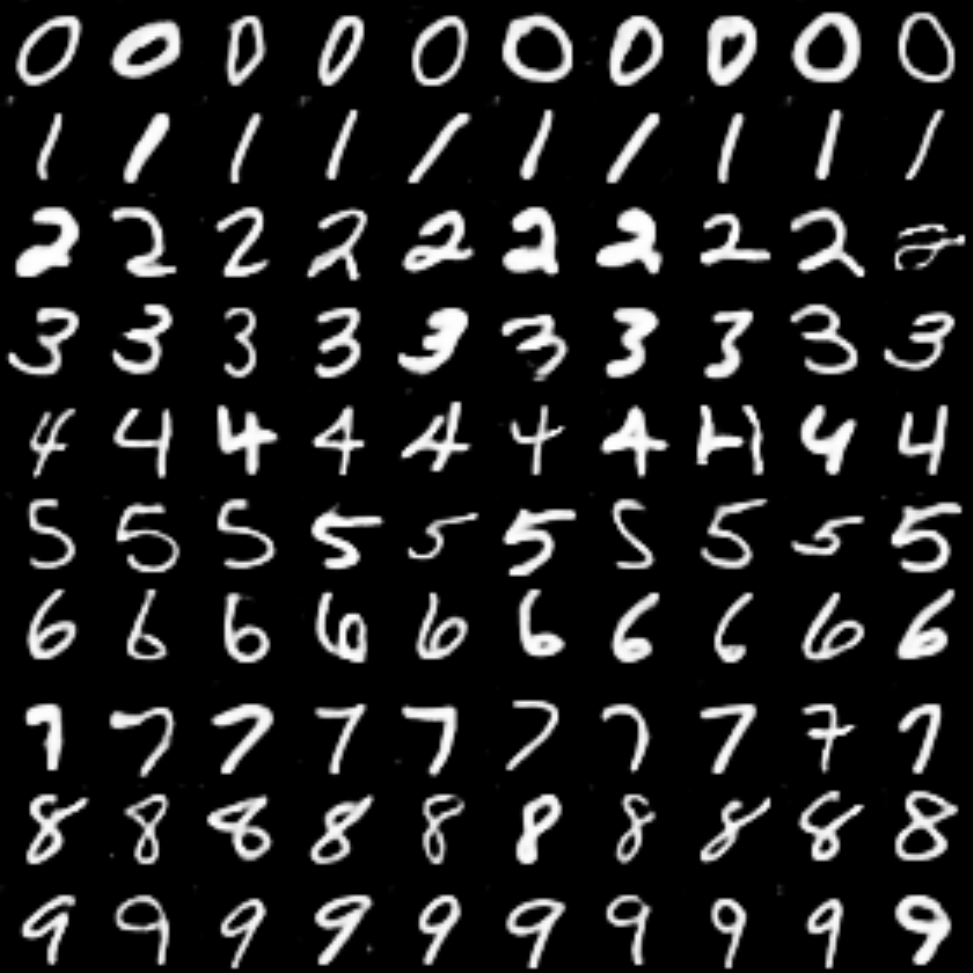}   \\
DCGAN & SteinGAN \\
\vspace{5\baselineskip}
\end{tabular}
\begin{comment}
\renewcommand{\arraystretch}{1.2}
\begin{tabular}{|l|c|c|c|}
\hline
Training Data & Real Images & DCGAN %& SteinGAN (w/o kernel) 
& SteinGAN \\
\hline
Testing Accuracy& 92.58 \% &  41.82 \%  %& 62.93 \% 
& 58.69 \%\\
\hline
\end{tabular}
\end{comment}
\caption{
 MNIST images generated by DCGAN and our SteinGAN. We use the joint model in \eqref{equ:pxy} to allow us to generate images for each digit. We set $m = 0.2$. 
%Upper: MNIST images generated by DCGAN and our SteinGAN conditional on each digit. We use the joint model in \eqref{equ:pxy} to allow us generate images for each digit. Lower: Testing accuracy when using simulated images (of the same size as the real training set) to train ResNets for classification. SteinGAN achieves higher testing accuracy than DCGAN.
}
\label{fig:mnist}
\end{figure}

\paragraph{Discussion}
The MNIST dataset has a training set of $60,000$ examples. 
% Our discriminator has 3 layers for both encoder and decoder with shared parameters. 
Both DCGAN and our model produce high quality images, both visually indistinguishable from real images; see figure \ref{fig:mnist}. 

CIFAR-10 is very diverse, and with only 50,000 training examples.
Figure~\ref{fig:cifar10} shows examples of simulated images by DCGAN and SteinGAN generated conditional on each category, which look equally well visually. 
%We provide two quantitive evaluation: 1) a recently proposed inception score \citep{salimans2016improved},
%we also provide quantitively evaluation using a recently proposed inception score \citep{salimans2016improved} 
We also provide quantitively evaluation using a recently proposed inception score \citep{salimans2016improved}, as well as 
the classification accuracy when training ResNet using $50,000$ simulated images as train sets, evaluated on a separate held-out testing set never seen by the GAN models.  
Besides DCGAN and SteinGAN, we also evaluate another simple baseline obtained by subsampling 500 real images from the training set and duplicating them 100 times. 
We observe that these scores capture rather different perspectives of image generation:
The inception score favors images that look realistic individually and have uniformly distributed labels; as a result, 
the inception score of the duplicated 500 images is almost as high as the real training set. 
We find that the inception score of SteinGAN is comparable, or slightly lower than that of DCGAN. 
On the other hand, the classification accuracy measures the amount information captured in the simulated image sets;  
% measures the realisiticity of each image 
%Interestingly, we find the inception score of the duplicated 500 images is almost as high as the real training set. 
%On the other hand, the testing accuracy captures the amount of information included in the image set.  
we find that SteinGAN achieves the highest classification accuracy, suggesting that it captures more information in the training set. %on the held-out testing set the model never saw. 
%We first evaluate a recently proposed inception score \cite{salimans2016improved} for all models. Figure \ref{fig:cifar10} shows the generated examples as well as the inception score. However, we find by sampling choosing 500 images and duplicating them 100 times, we could get an inception score almost as high as on the real training set. We propose to use the samples to train a classifier and test on the real testing images, which, none of the GAN models saw before during the learning stage. Our SteinGAN achieves higher accuracy.

\begin{figure}[t]
\centering
\begin{tabular}{cc}
\includegraphics[width=0.45\textwidth]{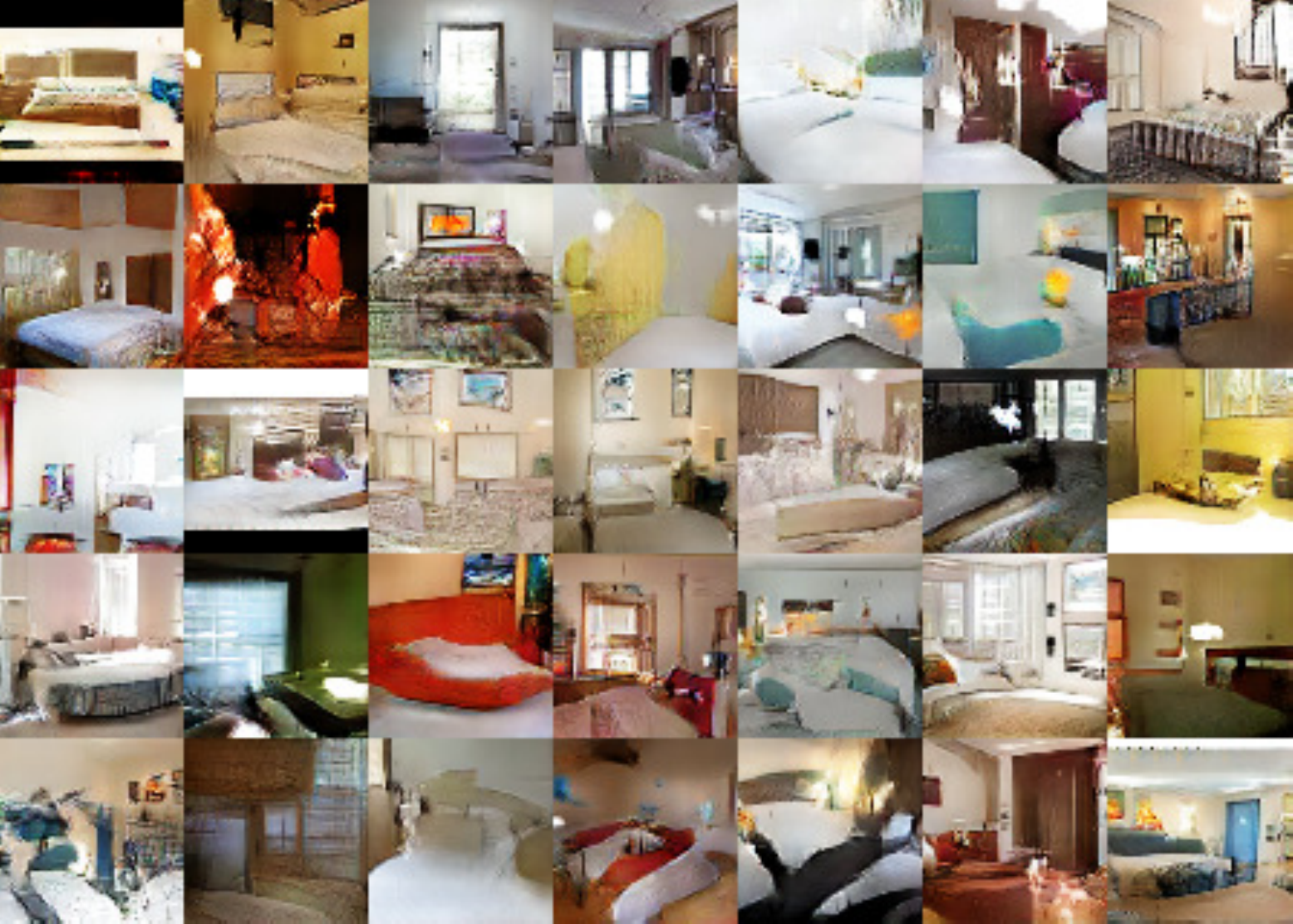} & 
\includegraphics[width=0.45\textwidth]{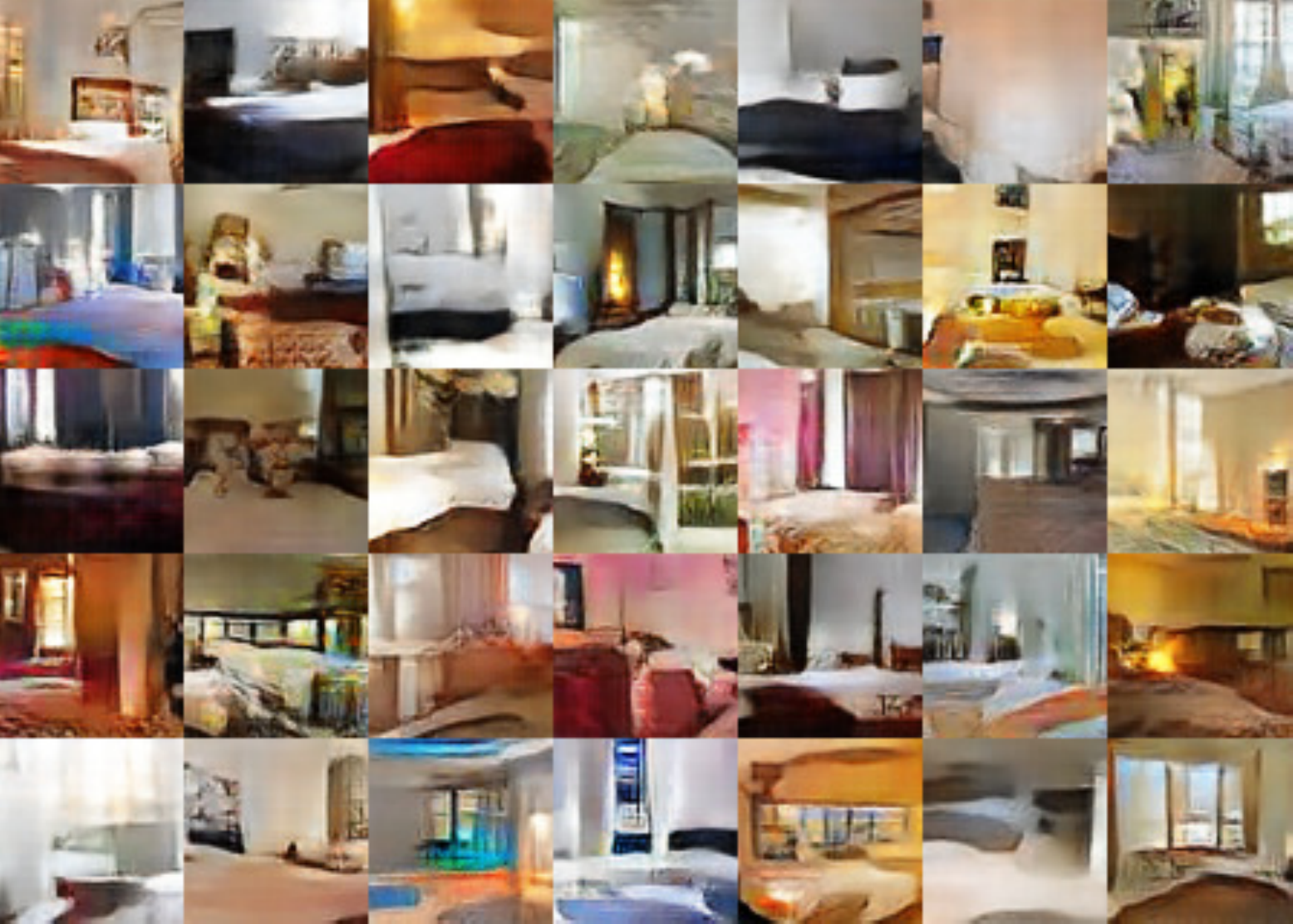} \\
DCGAN & SteinGAN\\
\end{tabular}
\caption{Images generated by DCGAN and our SteinGAN on LSUN.}
\label{fig:room}
\end{figure}

Figure~\ref{fig:face} and \ref{fig:room} visualize the results on CelebA (with more than 200k face images) and LSUN (with nearly 3M bedroom images), respectively. 
We cropped and resized both dataset images into $64\times 64$. 
%CelebA has more than 200K face images, and the LSUN dataset has nearly 3M bedroom images. We cropped and resized both dataset images into $64\times 64$.
%The generated samples are visualized in figure \ref{fig:face} and \ref{fig:room}.

\begin{comment}
\begin{figure}[h]
\centering
\includegraphics[width=0.9\textwidth]{figures/faces/vgd_gan-20.pdf}  
\caption{More images generated by SteinGAN on CelebA.}
\label{fig:facemore}
\end{figure}

\end{comment}

\section{Conclusion}
We propose a new method to train neural samplers for given distributions, together with a new SteinGAN method for generative adversarial training. 
Future directions involve more applications and theoretical understandings for training neural samplers. 

\newpage\clearpage
\bibliographystyle{iclr2016_conference}
{\small
\bibliography{bibrkhs_stein}
}

%\newpage
\appendix

\end{document}